\documentclass[letterpaper, 10 pt, conference]{ieeeconf}  

\IEEEoverridecommandlockouts
\usepackage{graphicx}
\graphicspath{{figs/}}
\usepackage{amsmath}
\usepackage{amssymb}
\usepackage{algorithm}
\usepackage{algpseudocode}
\usepackage{times}
\usepackage{url}

\title{\LARGE \bf
A Face Robot which Imitates Human Musculoskeletal Skin System \\
and Its Evaluation on Basic Movement of Facial Expression
}

\author{Shun Kaneko$^{1}$, Shunnosuke Yoshimura$^{1}$, Akihiro Miki$^{1}$, Kento Kawaharazuka$^{1,2}$, Kei Okada$^{1}$
\thanks{$^{1}$ The authors are with the Department of Mechano-Informatics, Graduate School of Information Science and Technology, The University of Tokyo, 7-3-1 Hongo, Bunkyo-ku, Tokyo, 113-8656, Japan.
    {\texttt\small [kaneko, yoshimura, miki, kawaharazuka, k-okada]@jsk.imi.i.u-tokyo.ac.jp}
  }
    \thanks{$^{2}$ The author is with the AI Center, Graduate School of Information Science and Technology, The University of Tokyo, Japan.
  }
}

\begin{document}
\maketitle
\thispagestyle{empty}
\pagestyle{empty}
\begin{abstract}
Facial robots require many coordinated, small-amplitude motions, but accommodating these motions in a compact face remains difficult.
This paper presents an anatomically inspired facial-robot platform that reproduces the layered organization of the human facial musculoskeletal system.
The platform comprises bone, muscle, and skin components whose geometry, shape, and attachment locations are derived from three-dimensional anatomical data.
Wire-driven, compliant muscle components are routed through PTFE tubes, allowing actuation to be placed away from the face.
We describe the selection and fabrication of 65 muscles (34 muscle types) required to reproduce the 47 target action units (AUs) in the Facial Action Coding System (FACS).
We then evaluate the reproducibility of individual AUs in two stages: muscle actuation without skin and actuation after the silicone skin is attached.
Forty of the 47 target AUs were reproduced with the muscle components, and 30 were reproduced after skin attachment.
These results demonstrate the feasibility of reproducing a broad range of FACS facial expressions using an anatomically inspired musculoskeletal structure.
The proposed platform provides a foundation for the systematic development and evaluation of future anatomically inspired facial robots.

\end{abstract}
\section{INTRODUCTION}

\begin{figure}[thpb]
  \centering
  \includegraphics[width=\columnwidth]{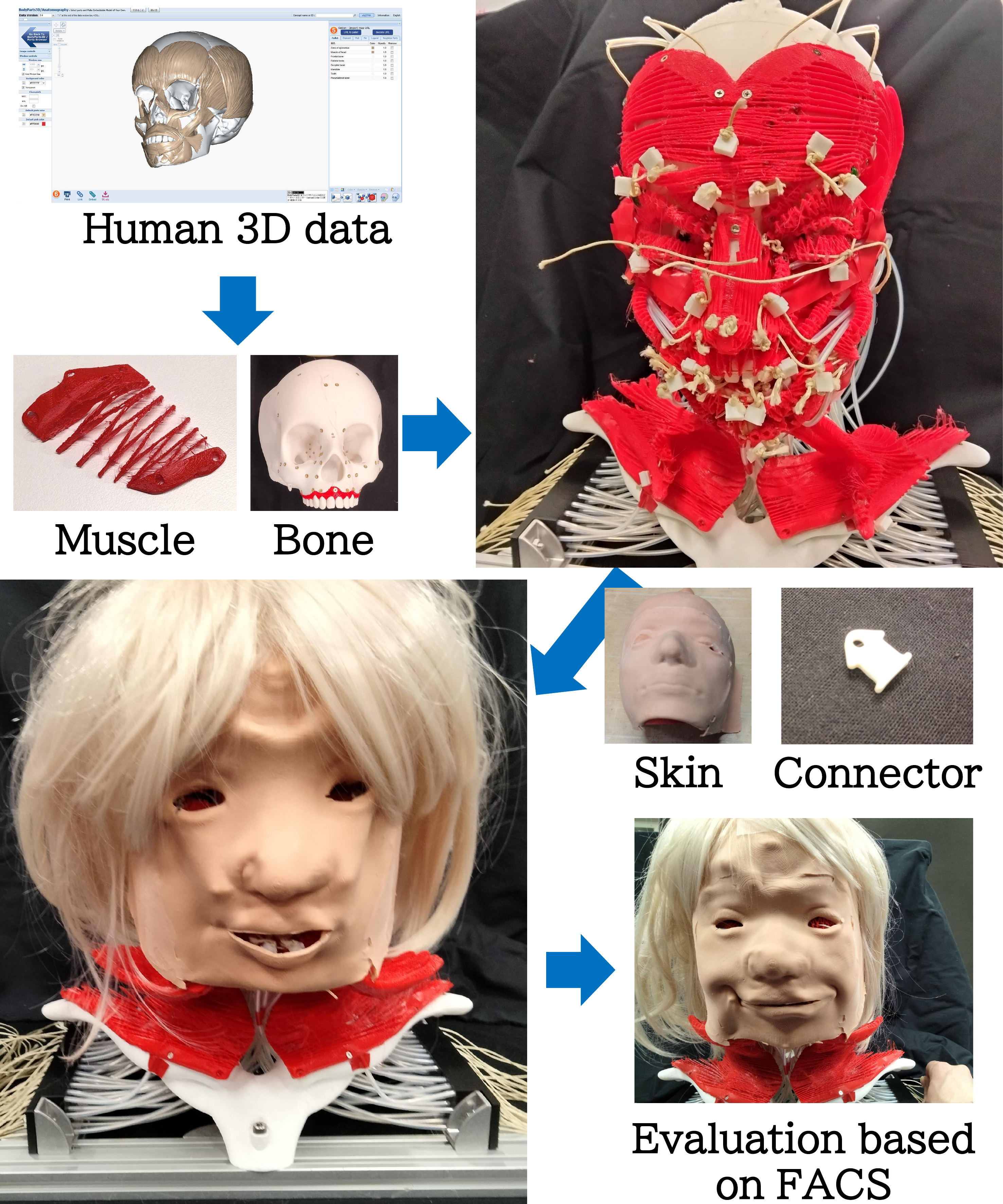}
  \caption{Overview of the proposed anatomically inspired facial-robot platform. Bone, muscle, and skin components are derived from anatomical 3D data. We assess the reachability of individual action units (AUs) of facial action coding system (FACS) \cite{FACS}, first with the muscle components and then with the skin attached.}
  \label{figure:robot_overview}
\end{figure}

Facial expressions play a central role in communication between humans and humanoid robots. 
While many humanoid robots incorporate facial expressions, reproducing numerous small, coordinated facial motions within the limited space of a human-sized face remains a substantial mechanical challenge. 
Facial robots therefore span a wide spectrum, ranging from platforms with few or no expressive facial features \cite{UnitreeG1,Atlas} to highly expressive androids capable of generating detailed facial motions \cite{ameca,erica}.

Existing expressive facial robots can be broadly categorized according to their actuation and structural principles. 
Linkage-driven and servomotor-driven systems, such as SAYA \cite{saya} and Ameca \cite{ameca}, mechanically transmit actuator motion to the face through motors, links, and joints. 
Although these mechanisms provide precise control, the need to package rigid components within a face-sized volume limits the number and placement of independently actuated facial motions. 
Another approach is to design the robot according to the biological facial musculoskeletal system, representing the face as an integrated structure of bones, muscles, and skin. 
Musculoskeletal robots have successfully demonstrated anatomically inspired actuation for body movements \cite{Yoshimura,mckibben}; 
however, applying this design principle systematically to an entire facial structure capable of reproducing diverse facial expressions remains comparatively unexplored.

In this work, we present an anatomically inspired facial-robot platform that reproduces the layered organization of the human facial musculoskeletal system. 
The platform comprises of anatomically derived bone, muscle, and skin components whose geometries and attachment locations are based on BodyParts3D anatomical data \cite{bodyparts3d}. 
Facial muscles are implemented as compliant wire-driven components actuated through PTFE tubes, allowing the actuators to be located away from the face while preserving anatomically realistic muscle paths.

The objective of this study is to present the design of this anatomically inspired facial-robot platform and to evaluate its capability to reproduce facial expressions defined by the Facial Action Coding System (FACS) \cite{FACS}. 
FACS is used both to determine the muscles required for facial expression reproduction and to assess the reproducibility of individual Action Units (AUs). 
The evaluation is conducted in two stages: first with the muscle components alone and then after attaching the silicone skin. 
Of the 47 target AUs, 40 were reproduced by the muscle components and 30 remained reproducible after skin attachment.

The main contributions of this paper are as follows:
(1) the anatomical design and fabrication of a facial robot comprising 65 wire-driven muscle components, anatomically derived bone structures, and silicone skin;
(2) the implementation of an anatomically inspired facial musculoskeletal system designed to reproduce all 47 target Action Units (AUs) defined in FACS; and 
(3) an experimental evaluation of the extent to which an anatomically inspired facial musculoskeletal structure can reproduce AUs defined by FACS.

\section{SELECTION OF MUSCLES TO BE IMITATED}

\begin{figure}[thpb]
  \centering
  \includegraphics[width=\columnwidth]{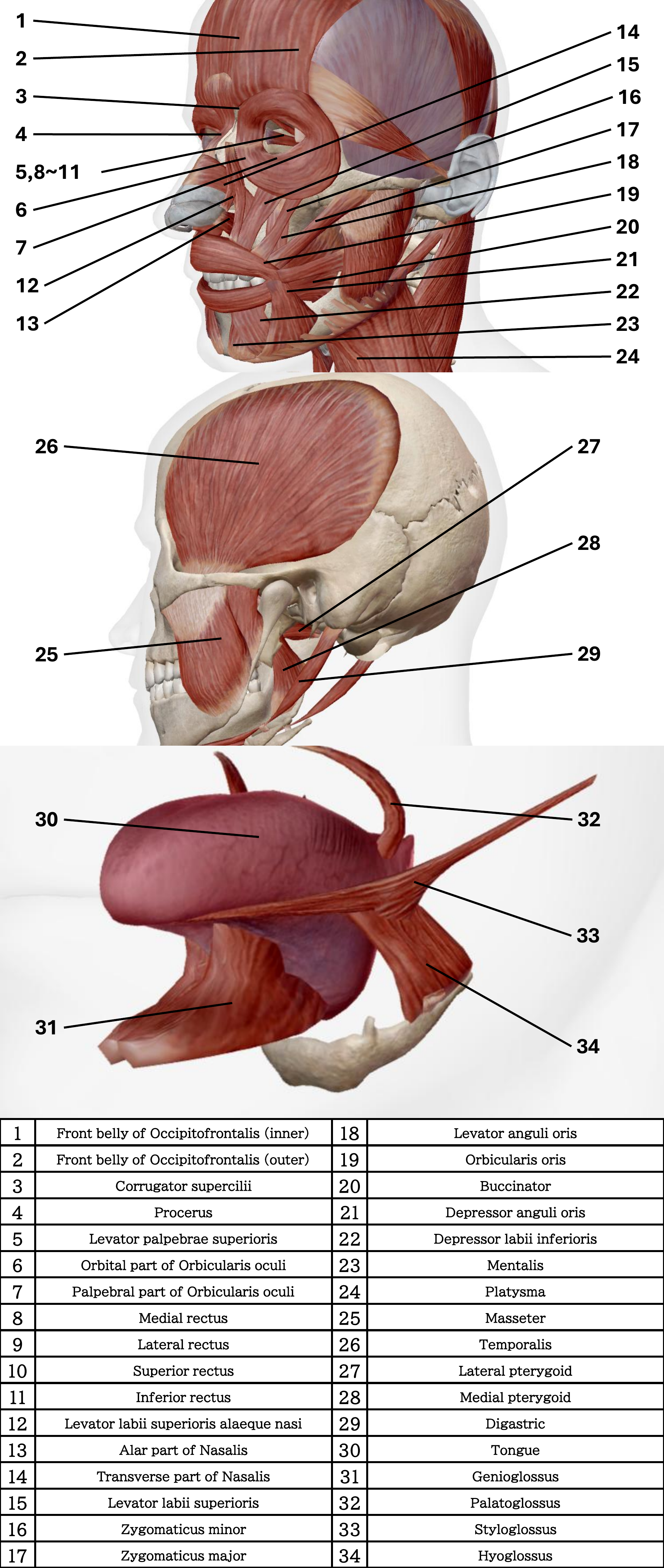}
  \caption{Muscles selected for the proposed platform. The selection is based on FACS \cite{FACS}. Covering the 47 target AUs requires 34 muscle types, totaling 65 individual muscles. The illustration was prepared using Visible Body Suite \cite{visiblebodysuite}; muscle names follow Ito \cite{muscle_anatomy}.}
  \label{figure:muscles}
\end{figure}

To define both the anatomical scope of the proposed platform and its evaluation target, we adopt the Facial Action Coding System (FACS) \cite{FACS}. 
FACS describes facial movements as combinations of Action Units (AUs), each corresponding to the visible effect of one or more facial muscles. 
Because it provides a standardized relationship between facial muscle activity and observable facial motion, 
FACS serves as a standardized framework for both selecting the muscles to be implemented and evaluating the facial motions generated by the proposed platform.

FACS has also been widely adopted to characterize the expressive capabilities of facial robots \cite{facs_apply_robot_1,facs_apply_robot_2}. 
Consistent with this practice, we use FACS to systematically assess the reproducibility of individual AUs by the proposed platform. 
The present evaluation focuses on the reproducibility of individual AUs, providing a foundation for more detailed quantitative and perceptual analyses in future studies.

The proposed platform targets the 47 principal AUs listed on the FACS manual score sheet \cite{FACS}. 
Based on the muscle–AU correspondence specified in FACS, we identified the facial muscles required to reproduce these target AUs. 
The resulting design comprises 34 muscle types and 65 individual muscles, as illustrated in Fig.~\ref{figure:muscles}. 
Targeting this broad set of AUs enables a comprehensive assessment of the facial expression capability of the proposed platform.
\section{DESIGN OF THE MUSCULOSKELETAL SYSTEM-IMITATING FACE ROBOT}

\subsection{Bone Components}

\begin{table}[h]
  \vspace{2pt}
  \caption{List of bone components}
  \label{table:bone_list}
  \begin{center}
    \begin{tabular}{p{1.5cm}|p{6.0cm}}
      \hline\hline
      component & bone \\
      \hline\hline
      Head & Maxillary teeth, maxillary gingiva, nasal bone, maxillae, zygomatic bone, temporal bone, parietal bone, occipital bone, sphenoid bone, frontal bone \\
      \hline
      Jaw & Mandibular teeth, mandibular gingiva, mandible \\
      \hline
      Tongue & Hyoid bone \\
      \hline
      Neck & C1–C7 cervical vertebrae, C2–C6 intervertebral discs \\
      \hline
      Shoulder & Manubrium of the sternum, first costal cartilage, first rib, clavicle, T1 thoracic vertebra, C7/T1 intervertebral disc \\
      \hline\hline
    \end{tabular}
  \end{center}
\end{table}

\begin{figure}[thpb]
  \centering
  \includegraphics[width=\columnwidth]{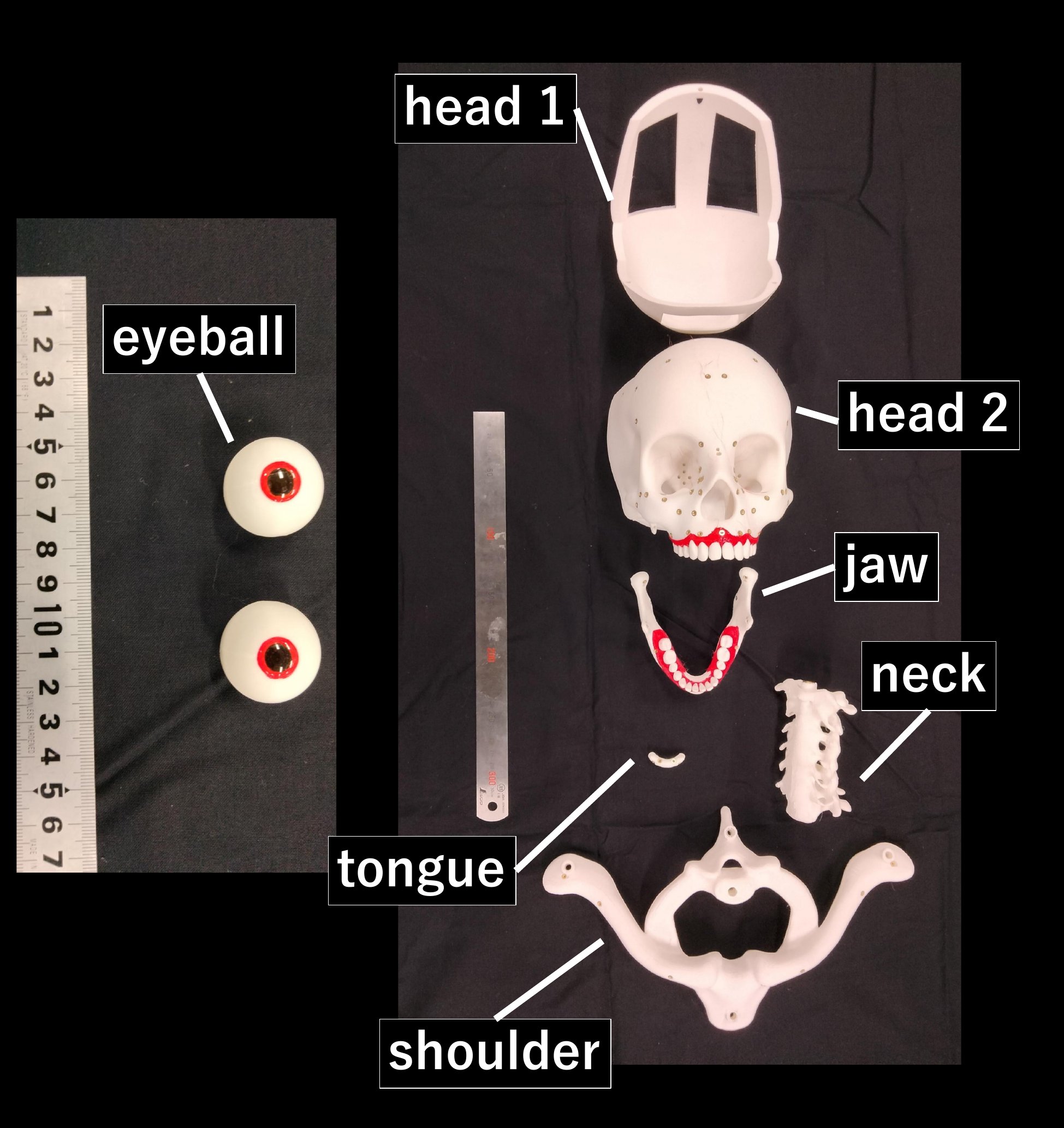}
  \caption{Photographs of the bone and eyeball components.
  The upper image shows the five bone assemblies used to simplify fabrication: head, jaw, neck, shoulder, and tongue. 
  The head is split in half for access to its interior. 
  The lower image depicts the eyeball parts; the pupils were drawn with a marker and sealed with clear resin for realism.}
  \label{figure:bone_components}
\end{figure}

The bone components provide the underlying structure for the robot. 
They were modelled from actual human 3D anatomical data 
but consolidated into five assemblies (head, jaw, neck, shoulder and tongue) to reduce complexity and make manufacturing easier.  
Table. \ref{table:bone_list} lists the elements included in each assembly. 
Although the maxillary and mandibular gingiva are soft tissues, we include them here for convenience.

Most pieces are printed in PLA, a common 3D-printing material with adequate strength and durability. 
Flexible sections such as the gingiva are printed in TPU. 
The neck, tongue and shoulder parts were produced on a Bambu Lab X1 Carbon printer, 
while the jaw and head were made on an Original Prusa XL printer.

Mechanical connections are formed with bolts between the neck and shoulder and between the head and neck. 
In contrast, the jaw, tongue and head are coupled by muscle components. 
Each bone part contains holes for insert nuts at the attachment points for muscles. 
The head is split into two halves to permit access to its interior for assembly and maintenance.

Although not skeletal, we will mention the eyeballs in this section. 
These are also printed in PLA. The pupils are then hand drawn and coated with clear resin. 
The eyeballs are mounted in the head and actuated by the muscle components. 

The final product is shown in Fig. \ref{figure:bone_components}.

\subsection{Muscle Components}

\begin{figure}[thpb]
  \centering
  \includegraphics[width=\columnwidth]{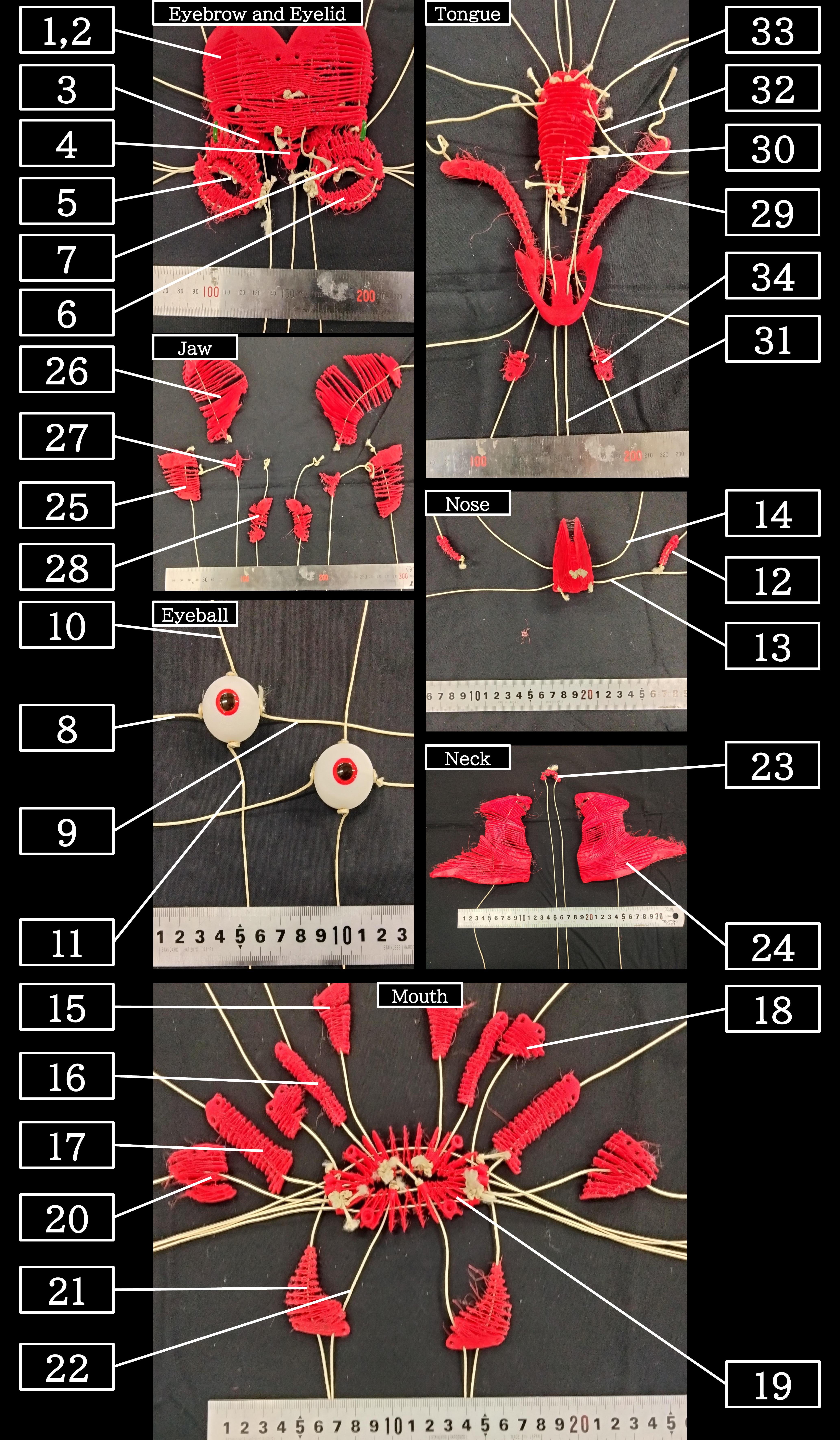}
  \caption{All muscle components. 
  Numbers match those in Fig. \ref{figure:muscles}. 
  Each unit comprises a fixing part (for attachment) and a zigzag contracting part.}
  \label{figure:muscle_components}
\end{figure}

The muscle components were designed to satisfy three primary requirements. 
First, the muscle components should be able to attach firmly to the bone components. 
Second, the muscle components should allow wires to pass through, enabling the wires to contract the muscle components. 
Finally, in addition to meeting the above two criteria, the muscle components should contract as much as possible.

\begin{figure}[thpb]
    \centering
    \includegraphics[width=\columnwidth]{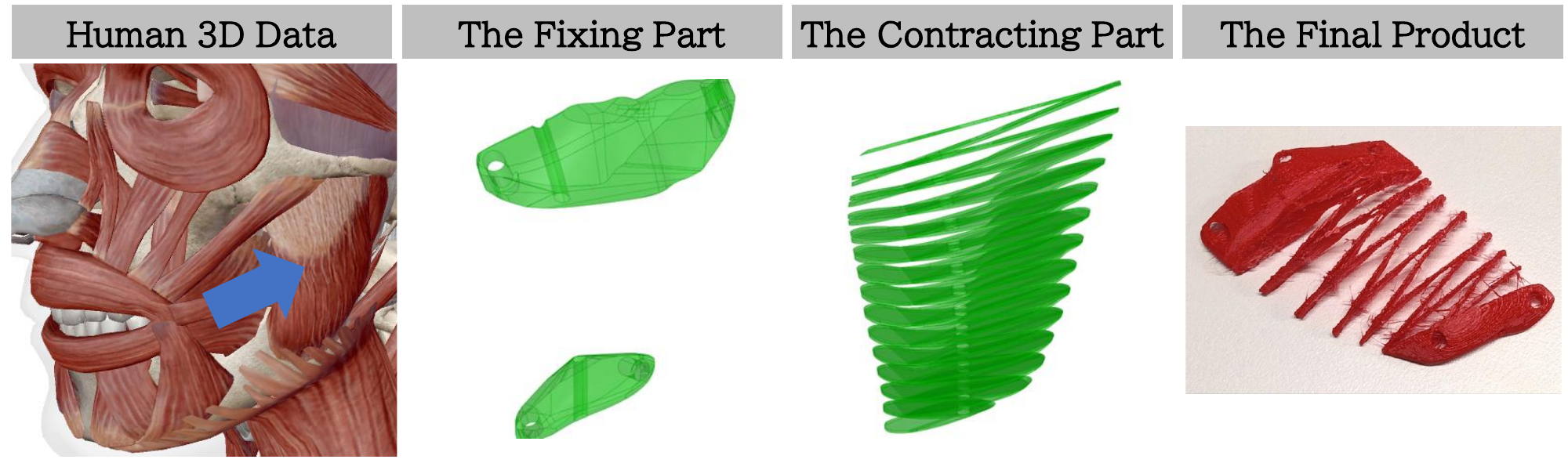}
    \caption{Overview of making muscle components. 
    From human 3D data, we create fixing parts and contracting parts. 
    We combine those parts to create the final product. }
    \label{figure:muscle_making_overview}
    \vspace{-15px}
\end{figure}

To meet these criteria, we designed the muscle components that consist of two subparts: the fixing part and the contracting part.
The fixing parts have bolt holes for attachment and guides for the wire for keeping the wire in place. 
The bolt holes must be strong enough to withstand the forces of muscle contraction. 
The contracting part is made of a thin, zigzag-shaped plate with a wire hole. 
Facial muscles are generally thin and flat, so the wire path should be as close to the center of the muscle component as possible.
The overall designing process is shown in Fig. \ref{figure:muscle_making_overview}

Designing the subparts separately allows us to optimize each for its function. 
All geometries were generated in Rhino using its Python scripting interface, enabling easy parameter changes. 
The design process for each subpart is detailed below.

\begin{itemize}
  \item
  \textbf{The Fixing Part}  
  
  This subcomponent must resist the forces generated during contraction while occupying minimal space to leave room for the contracting part. 
  Fig. \ref{figure:muscle_fixing_part} illustrates the procedure. 
  Bolt and wire path data are imported into Rhino, then wrapped with meshes at a distance of $X$. 
  These meshes are grouped and joined to form the basic shape of the fixing part. 
  The shape is intersected with the muscle 3D data, and the bolt and wire paths are hollowed out. 
  This procedure guarantees at least a thickness of $X$ around the holes while keeping the overall volume small. 
  The variable $X$ controls the strength of the fixing part.

  \begin{figure}[thpb]
    \vspace{15px}
    \centering
    \includegraphics[width=\columnwidth]{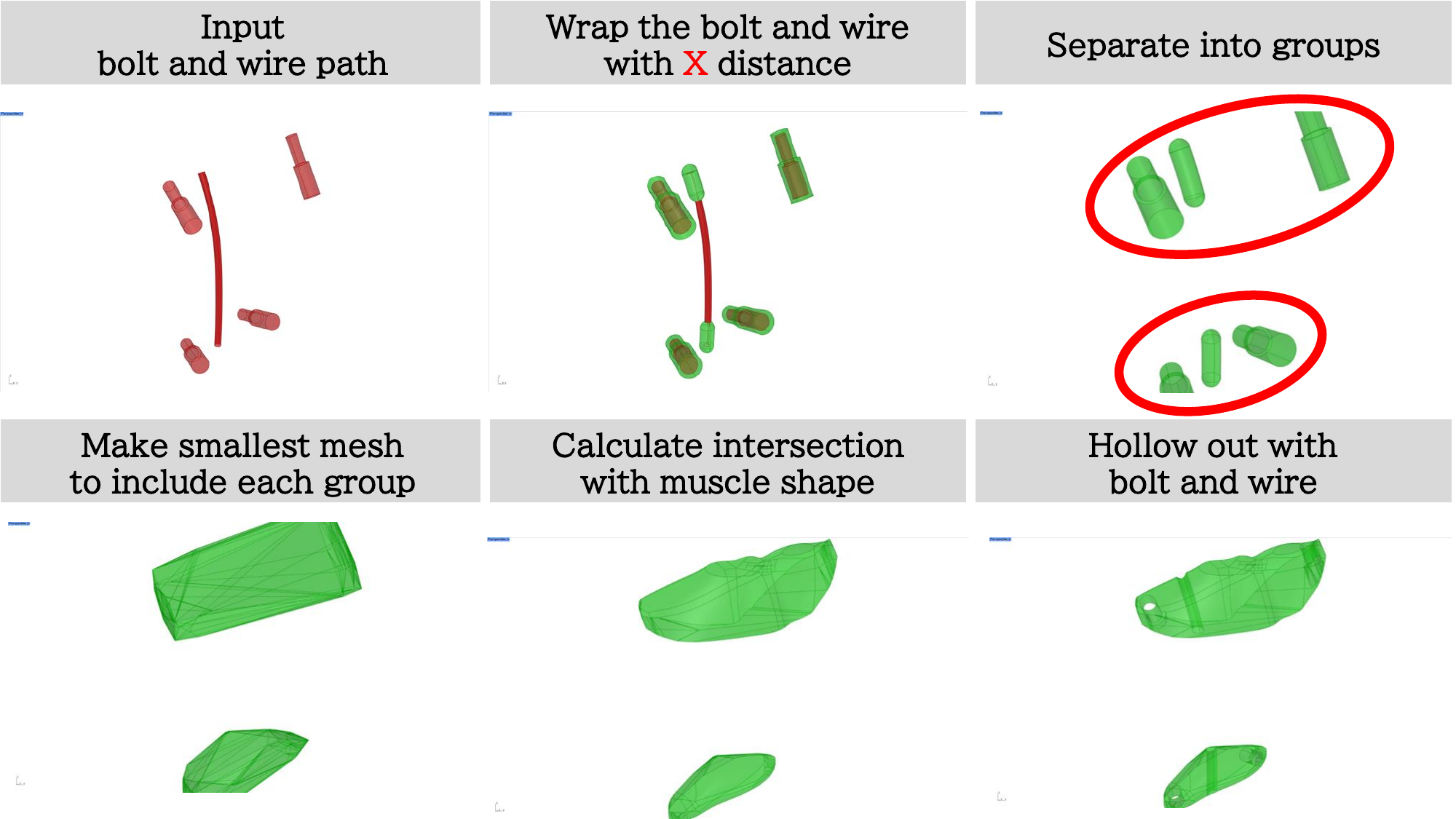}
    \caption{Process for generating a fixing part. 
    The method ensures a minimum thickness $X$ around bolt and wire holes while minimising the component's footprint.}
    \label{figure:muscle_fixing_part}
  \end{figure}

  \item
  \textbf{The Contracting Part} 
  
  To maximise contraction, the contracting part is shaped into a zigzag that follows the wire path. 
  The workflow is shown in Fig. \ref{figure:muscle_contracting_part}. 
  Muscle shape and wire paths are imported into Rhino; lines offset by a distance $X$ are drawn sequentially from the wire path. 
  Intersection points of these lines with the muscle surface are computed and connected to form a zigzag polyline. 
  This polyline is then extruded to a plate of thickness $Y$, and the resulting geometry is clipped by the muscle boundary. 
  The output is $X$ plates of thickness $Y$ arranged zigzag. 
  Increasing $Y$ raises stiffness at the cost of softness; reducing the number of plates $X$ increases achievable contraction but may reduce the part’s ability to retain its intended shape.

  \begin{figure}[thpb]
    \centering
    \includegraphics[width=\columnwidth]{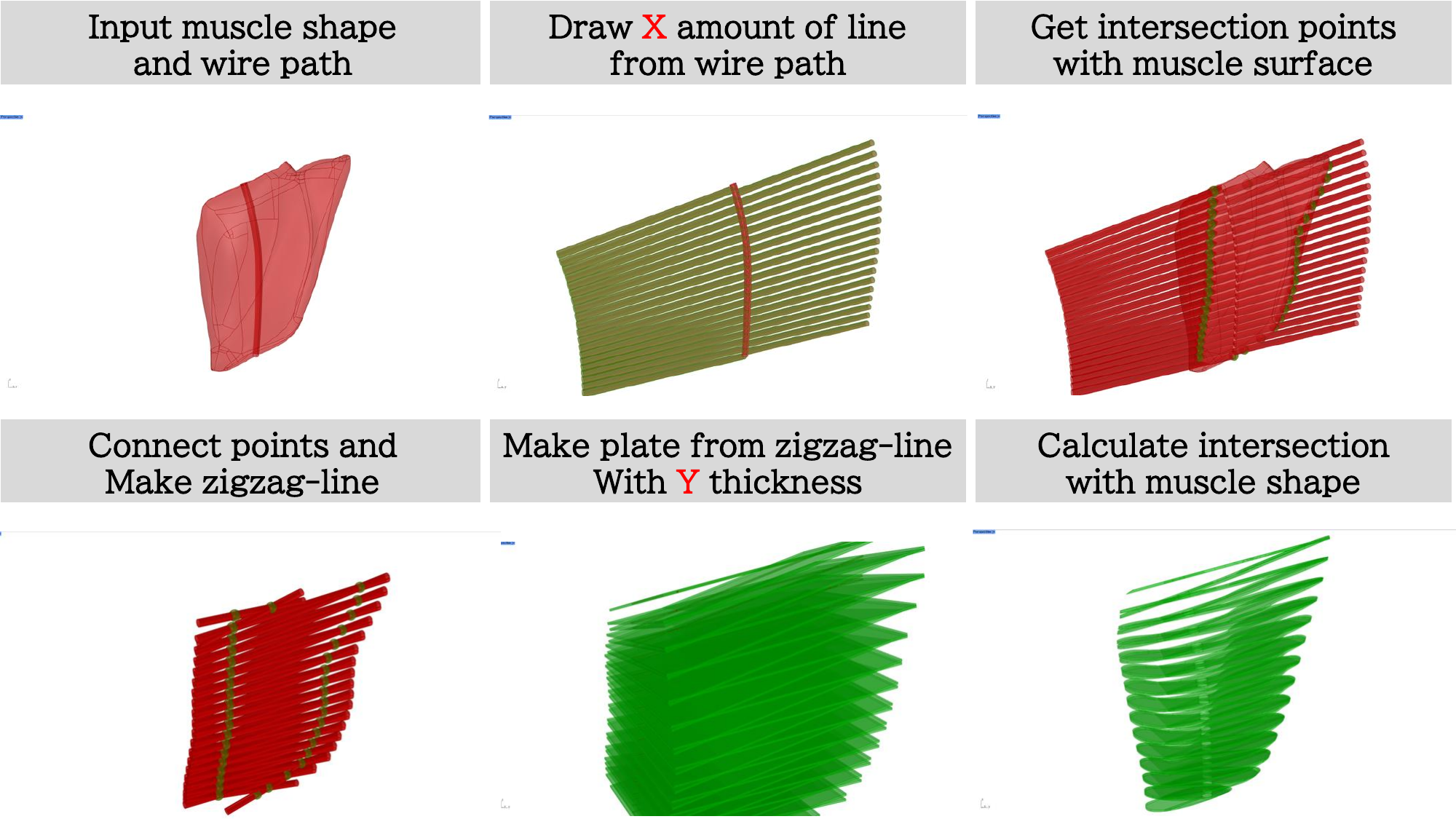}
    \caption{Workflow for creating a contracting part.  
    The process yields $X$ zigzag plates of thickness $Y$. 
    Adjusting $X$ and $Y$ trades off contraction range, softness, and structural integrity.}
    \label{figure:muscle_contracting_part}
  \end{figure}

\end{itemize}

The completed muscle components are printed in TPU on the Original Prusa XL printer. 
Fig. \ref{figure:muscle_components} shows all of the muscles used in our robot; the numbering corresponds to the muscles depicted in Fig. \ref{figure:muscles}. 

\subsection{Skin Components}

\begin{figure}[thpb]
  \centering
  \includegraphics[width=\columnwidth]{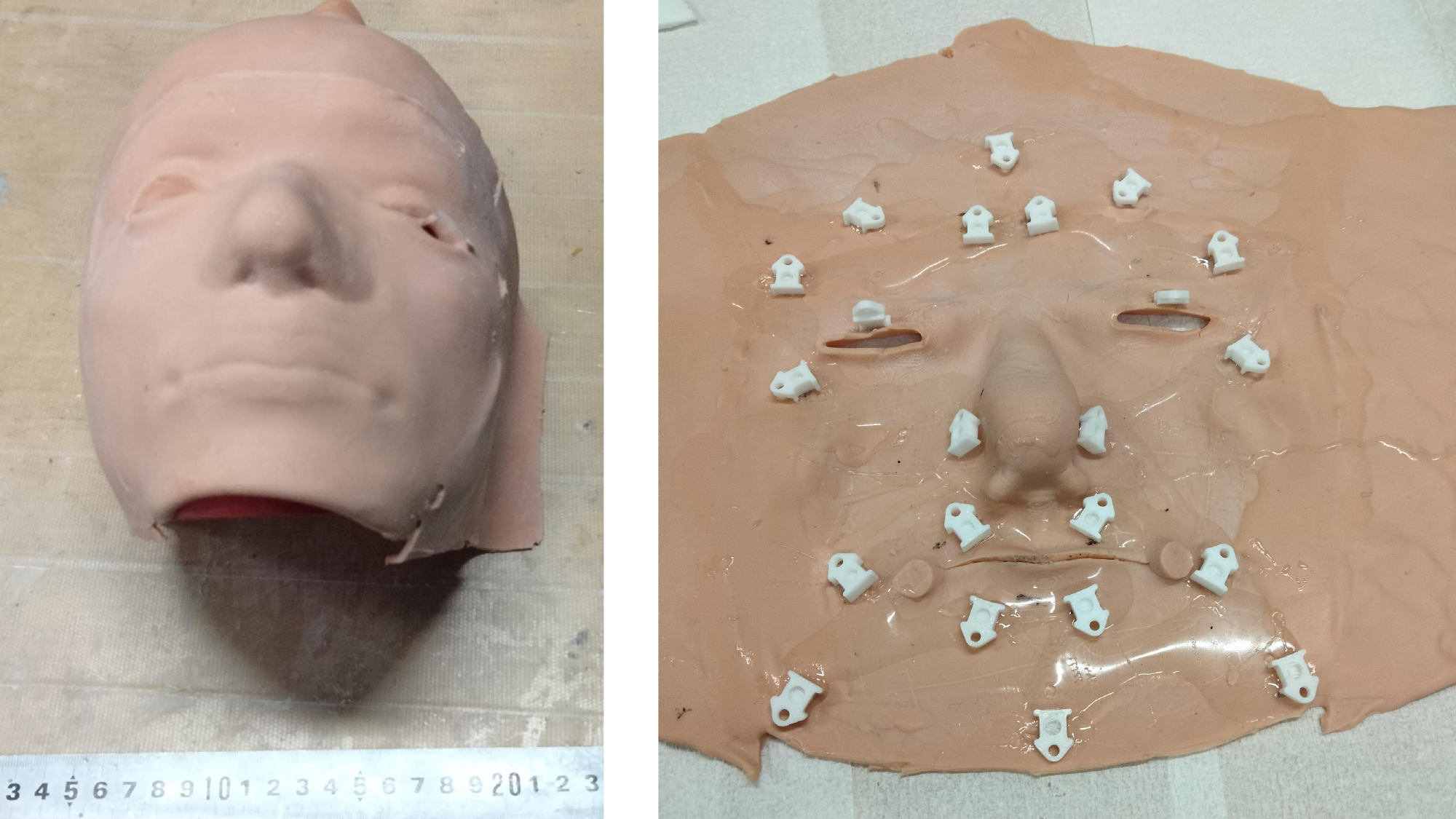}
  \caption{Skin components. 
  The parts are cast in silicone. 
  Because the molds have concave and convex features, each skin piece must be produced in several
  sections corresponding to allowable demolding directions; the sections are later joined. 
  The left image shows a completed component; the right image indicates connector locations.}
  \label{figure:skin_components}
\end{figure}

The skin must be sufficiently soft and flexible to deform under muscle action.
Accordingly, we cast the skin components from silicone, a material commonly used for robot skins \cite{ecoflex_apply_robot_1,ecoflex_apply_robot_2}. 
Molds were produced on the Bambu Lab X1 Carbon printer. 
Due to the use of concave and convex molding, the skin components are molded separately and later connected.

The assembled skin is attached to the bone components with tape. 
Connections between the skin and muscles are made with small connectors that are glued to the skin and then tied to the muscle wires.

The final product is shown in Fig. \ref{figure:skin_components}.

\section{EXPERIMENTS}

\subsection{Evaluation Protocol and Scope}

To evaluate the expressive capability of the proposed platform, 
we assessed the reproducibility of each of the 47 target FACS Action Units (AUs) in two stages. 
First, the muscle components were evaluated without the silicone skin by manually actuating each actuation wire 
and observing whether the intended anatomical muscle motion was produced. 
Second, the silicone skin was attached and the same evaluation procedure was repeated to determine whether the corresponding visible facial motion was successfully reproduced. 
An AU was regarded as reproduced when the intended motion defined by FACS was clearly observed during the corresponding evaluation stage.

This two-stage evaluation separately assesses the capability of the anatomically inspired musculoskeletal structure and the capability of the complete facial platform after skin attachment. 
By comparing these two conditions, the evaluation clarifies how the addition of the skin influences the reproduction of individual facial motions and provides a systematic assessment of the expressive capability of the proposed platform.

\begin{table}[htbp]
  \caption{Summary of the two-stage reachability evaluation for the 47 target AUs. \\
  $\checkmark$: achieved with skin attachment \\
  *: only muscle movement reproduced}
  \label{table:au_summary}
  \begin{center}
    \begin{minipage}[t]{0.31\linewidth}
      \centering
      \scriptsize
      \begin{tabular}{c | c}
        \hline \hline
        AU & result \\
        \hline \hline
        AU01 & $\checkmark$ \\
        AU02 & $\checkmark$ \\
        AU04 & $\checkmark$ \\
        AU05 & $\checkmark$ \\
        AU06 & $\checkmark$ \\
        AU07 & $\checkmark$ \\
        AU08 & $\checkmark$ \\
        AU09 & $\checkmark$ \\
        AU10 & $\checkmark$ \\
        AU11 & $\checkmark$ \\
        AU12 & $\checkmark$ \\
        AU13 & $\checkmark$ \\
        AU14 & $\checkmark$ \\
        AU15 & $\checkmark$ \\
        AU16 & $\checkmark$ \\
        AU17 & $\checkmark$ \\
        AU18 & $\checkmark$ \\
        \hline \hline
      \end{tabular}
    \end{minipage}
    \hfill
    \begin{minipage}[t]{0.31\linewidth}
      \centering
      \scriptsize
      \begin{tabular}{c | c}
        \hline \hline
        AU & result \\
        \hline \hline
        AU19 & * \\
        AU20 & $\checkmark$ \\
        AU21 & * \\
        AU22 & $\checkmark$ \\
        AU23 & $\checkmark$ \\
        AU24 &  \\
        AU25 & $\checkmark$ \\
        AU26 & $\checkmark$ \\
        AU27 & * \\
        AU28 &  \\
        AU29 & * \\
        AU30 & * \\
        AU31 & $\checkmark$ \\
        AU32 & $\checkmark$ \\
        AU33 &  \\
        AU34 &  \\
        AU35 &  \\
        \hline \hline
      \end{tabular}
    \end{minipage}
    \hfill
    \begin{minipage}[t]{0.31\linewidth}
      \centering
      \scriptsize
      \begin{tabular}{c | c}
        \hline \hline
        AU & result \\
        \hline \hline
        AU36 & * \\
        AU37 & * \\
        AU38 &  \\
        AU39 &  \\
        AU43 & * \\
        AU45 & * \\
        AU46 & * \\
        AU61 & $\checkmark$ \\
        AU62 & $\checkmark$ \\
        AU63 & $\checkmark$ \\
        AU64 & $\checkmark$ \\
        AU65 & $\checkmark$ \\
        AU66 & $\checkmark$ \\
        \hline \hline
        $\checkmark$ total & 30 \\
        * total & 10 \\
        \hline \hline
      \end{tabular}
    \end{minipage}
  \end{center}
  \vspace{-15px}
\end{table}

Table~\ref{table:au_summary} summarizes the outcome. The two-stage design separates failures in muscle actuation from failures introduced by skin coupling or interactions in the assembled face.

\subsection{Evaluation of Muscle Movement}

\begin{figure}[thpb]
  \centering
  \includegraphics[width=\columnwidth]{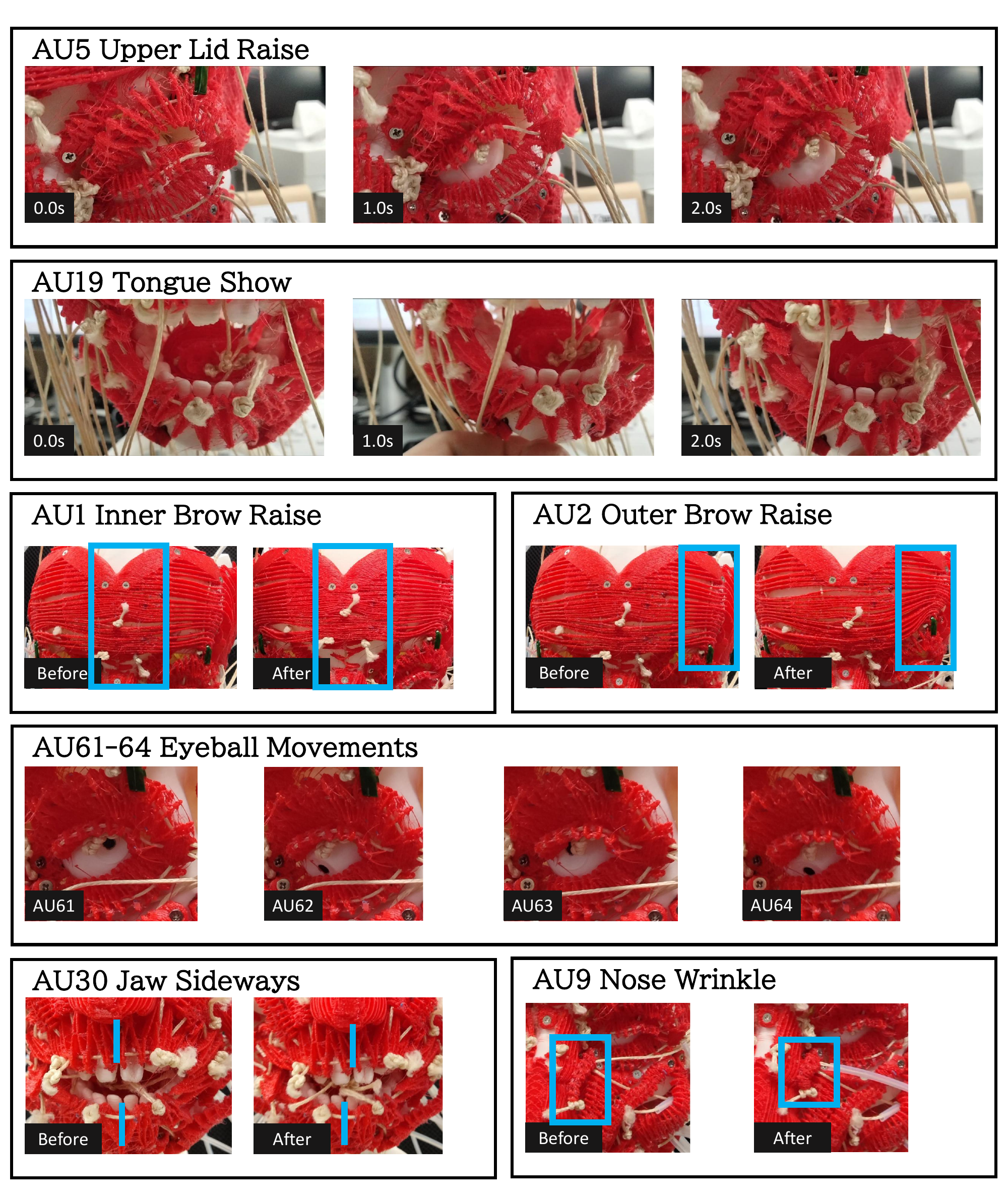}
  \caption{Example frames from the muscle-movement test. Wires were pulled to contract the muscle components and assess the intended individual AU motion. Forty of the 47 target AUs were observed at this stage.}
  \label{figure:muscle_movement_test}
\end{figure}

In the first stage, 40 of the 47 target FACS Action Units (AUs) were successfully reproduced by the muscle components without the silicone skin (Fig.~\ref{figure:muscle_movement_test}). 
This result demonstrates that the anatomically inspired musculoskeletal structure can reproduce the majority of the targeted facial motions through muscle actuation alone, confirming the feasibility of the design before considering the influence of the skin.

Representative examples are demonstrated across multiple facial regions. 
For example, AU1 (Inner Brow Raiser) and AU2 (Outer Brow Raiser) were reproduced by independently actuating the medial and lateral portions of the frontalis muscle, 
resulting in selective elevation of the inner and outer eyebrows, respectively. 
AU30 (Jaw Sideways) was achieved through coordinated actuation of the masticatory muscles, 
enabling lateral movement of the mandible. 
Eye movements (AU61--64) were reproduced by actuating the extraocular muscles surrounding the eyeballs, 
while AU19 (Tongue Show) was reproduced by protruding the tongue through coordinated actuation of the tongue muscles together with the mentalis muscle. 
These examples demonstrate that the proposed platform reproduces not only superficial facial deformations but also anatomically coordinated motions involving the eyebrows, jaw, eyeballs, and tongue.

The remaining seven AUs (AU24, AU28, AU33--35, and AU38--39) were not reproduced during this stage. 
These unreproduced AUs can be attributed to three implementation-specific factors. 
First, AU33--35 require changes in intraoral pressure for cheek puffing and sucking, whereas the current platform is driven solely by wire actuation. 
Second, AU38--39 depend on the nasalis muscle, whose anatomical attachment could not be accurately reproduced because the available anatomical data did not provide a suitable attachment configuration. 
Third, AU24 and AU28 require inward deformation of the lips, which could not be generated because the current zigzag TPU orbicularis oris component provides insufficient compliance in that direction.

\subsection{Evaluation with Skin Attachment}

\begin{figure}[thpb]
  \centering
  \includegraphics[width=\columnwidth]{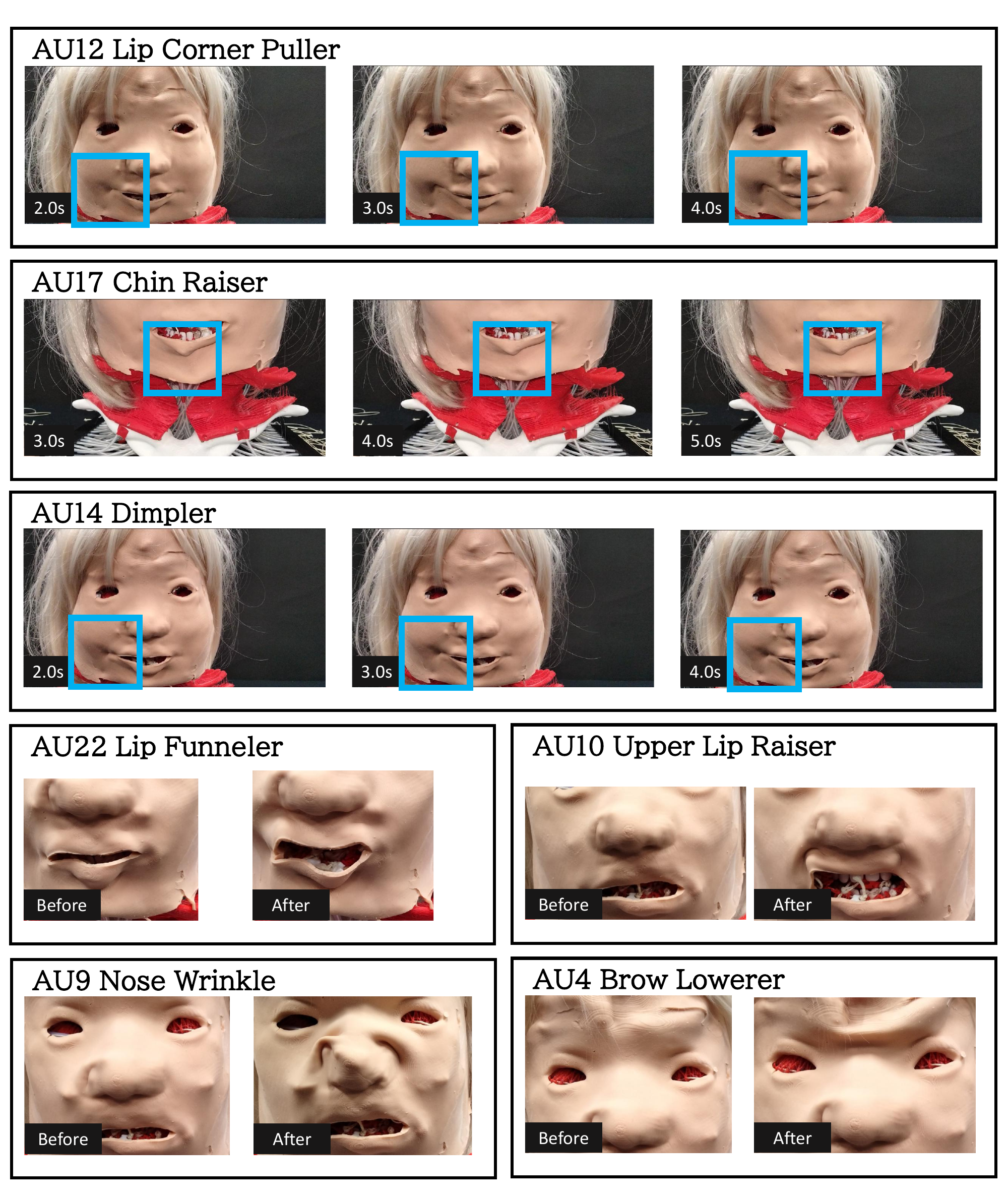}
  \caption{Example frames from the skin-attachment test. After attaching the skin, 30 of the 40 AUs observed in the muscle test remained visibly achievable. The images are qualitative examples, not displacement measurements.}
  \label{figure:skin_attached_test}
\end{figure}

After attaching the silicone skin, 30 of the 40 Action Units (AUs) that had been reproduced in the muscle-only evaluation remained visibly achievable (Fig.~\ref{figure:skin_attached_test}). 
These results demonstrate that the motions generated by the anatomically inspired musculoskeletal structure were successfully transferred to the skin, allowing a wide range of facial expressions to be reproduced by the assembled platform.

Representative anatomically realistic skin deformations were observed in several facial regions. 
In particular, AU17 (Chin Raiser) produced a distinct protrusion of the mental region. 
In the human face, contraction of the mentalis muscle elevates the lower lip by generating a characteristic bulging of the chin. 
A similar deformation was reproduced by the proposed platform, indicating that reconstructing the underlying musculoskeletal structure also reproduced the associated surface morphology. 
Likewise, AU10 (Upper Lip Raiser) generated a distinct nasolabial fold, a characteristic facial crease extending from the side of the nose toward the corner of the mouth. 
Similar anatomically realistic surface features were also observed in AU9 (Nose Wrinkler), which produced wrinkles over the nasal region, and AU12 (Lip Corner Puller), which generated wrinkles around the corners of the mouth. 
These examples demonstrate that the proposed anatomical design reproduces not only the intended facial motions but also characteristic skin deformations that naturally emerge from coordinated interactions between the muscle and skin components.

The remaining ten AUs (AU19, AU21, AU27, AU29--30, AU36--37, AU43, and AU45--46) were not visibly reproduced after skin attachment. 
These unreproduced AUs can be classified into four implementation-specific categories. 
First, AU43 and AU45--46 were affected by insufficient coupling between the orbicularis oculi component and the silicone skin, preventing eyelid closure. 
Second, AU19 and AU36--37 depend on tongue protrusion, but the tongue component derived from the anatomical data was too short to extend beyond the oral cavity. 
Third, AU27 and AU29--30 were restricted by mechanical interference between the jaw components and the routed wire tubes after assembly. 
Finally, AU21 requires deformation of the neck skin, which is outside the anatomical scope of the present platform.

\section{DISCUSSION}

The experimental results demonstrate that reproducing the anatomical organization of the human facial musculoskeletal system enables a broad range of facial motions and anatomically characteristic surface deformations. 
Unlike conventional linkage-driven facial robots, in which the skin is deformed directly by mechanical elements, the proposed platform generates facial motion through interactions among bone, muscle, and skin components. 
Representative examples include the chin protrusion generated during AU17 (Chin Raiser), the nasolabial fold produced during AU10 (Upper Lip Raiser), the nasal wrinkles generated in AU9 (Nose Wrinkler), and the wrinkles around the mouth produced in AU12 (Lip Corner Puller). 
These surface features emerge naturally from muscle contraction and skin deformation, suggesting that reproducing the underlying musculoskeletal structure provides a promising approach for generating anatomically realistic facial expressions.

The two-stage evaluation further distinguishes the capability of the musculoskeletal structure from that of the assembled facial platform. 
Before skin attachment, the developed muscle components reproduced 40 of the 47 target AUs, demonstrating that the anatomical muscle arrangement itself is capable of generating a wide variety of facial motions. 
After skin attachment, 30 AUs remained visibly reproducible, indicating that the layered musculoskeletal structure successfully transfers many of these motions to the facial surface while also revealing the influence of skin attachment on expression generation.

The unreproduced AUs were primarily associated with implementation-specific limitations rather than from the anatomical design concept itself. 
For example, cheek puffing requires pneumatic pressure that is not provided by the current wire-driven system. 
Nostril motion was limited by the available anatomical attachment data, while several jaw-related motions were affected by wire routing after assembly and tongue-related motions were constrained by the geometry of the current tongue component.
These observations suggest that the remaining limitations can be addressed through refinement of individual components while preserving the overall anatomical design philosophy.

Although the present evaluation focused on the reproducibility of individual FACS Action Units, more comprehensive evaluation remains necessary. 
Future work will include quantitative measurement of muscle contraction and skin deformation, objective comparison with human facial movements, evaluation of coordinated multi-AU expressions, and replacement of manual wire actuation with remotely driven actuators to enable repeatable experiments.

\section{CONCLUSION}

This paper presents an anatomically inspired facial-robot platform composed of bone, wire-driven muscle, and silicone skin components derived from human anatomical data. 
The proposed platform implements 65 individual muscles corresponding to the 47 principal FACS Action Units and provides a complete layered facial musculoskeletal structure.

A two-stage evaluation demonstrated that the proposed musculoskeletal structure reproduced 40 of the 47 target AUs without skin attachment and that 30 of these AUs remained visibly reproducible after the silicone skin was attached. 
In addition to reproducing a broad range of facial motions, the proposed platform generated anatomically realistic surface features, including chin protrusion, nasolabial folds, and facial wrinkles, through interactions between the muscle and skin components.

These results demonstrate that reproducing the human facial musculoskeletal system is a feasible approach for generating diverse facial motions together with the realistic skin deformations that accompany them. 
The proposed platform provides a foundation for future quantitative evaluation and further development of anatomically inspired facial robots.

\section*{ACKNOWLEDGEMENT}
This work was partially supported by JST Moonshot R\&D Program Japan Grant Number JPMJMS263D.
\addtolength{\textheight}{-12cm}

\end{document}